\documentclass[10pt,twocolumn,letterpaper]{article}

\usepackage[square,numbers]{natbib}
\usepackage{cvpr}
\usepackage{xcolor,xspace}
\usepackage{algorithm}
\usepackage[noend]{algpseudocode}
\usepackage{amsmath}
\usepackage{amssymb}
\usepackage{booktabs}
\usepackage{balance}
\usepackage{float}
\usepackage{amsfonts}      
\usepackage{url}
\renewcommand{\etal}{\emph{et al.}}
\newcommand{\comment}[1]{}

\renewcommand{\ie}{\emph{i.e.},~}
\renewcommand{\vs}{\emph{vs.~}}
\usepackage{times}
\usepackage{epsfig}
\usepackage{graphicx}
\newtheorem{prop}{Proposition}

\newcommand{\pframe}{\textit{P-frame}\xspace}
\newcommand{\iframe}{\textit{I-frame}\xspace}
\newcommand{\pframes}{\textit{P-frames}\xspace}
\newcommand{\iframes}{\textit{I-frames}\xspace}
\newcommand{\bframes}{\textit{B-frames}\xspace}

\usepackage[pagebackref=true,breaklinks=true,letterpaper=true,colorlinks,bookmarks=false]{hyperref}

\cvprfinalcopy 

\def\cvprPaperID{8215} 

\ifcvprfinal\fi
\begin{document}

\title{Mimic The Raw Domain: Accelerating Action Recognition in the Compressed Domain}
\author{Barak Battash\\
Intel\\
Haifa, Israel\\
{\tt\small barak.battach@intel.com}
\and
Haim Barad\\
Intel\\
Haifa, Israel\\
{\tt\small haim.barad@intel.com}
\and
Hanlin Tang\\
Intel Labs\\
San Francisco, CA\\
{\tt\small hanlin.tang@intel.com}
\and
Amit Bleiweiss\\
Intel\\
Haifa, Israel\\
{\tt\small amit.bleiweiss@intel.com}
}

\maketitle
\thispagestyle{empty}

\begin{abstract}
       Video understanding usually requires expensive computation that prohibits its deployment, yet videos contain significant spatiotemporal redundancy that can be exploited. In particular, operating directly on the motion vectors and residuals in the compressed video domain can significantly accelerate the compute, by not using the raw videos which demand colossal storage capacity. Existing methods approach this task as a multiple modalities problem. In this paper we are approaching the task in a completely different way; we are looking at the data from the compressed stream as a one unit clip and propose that the residual frames can replace the original RGB frames from the raw domain. Furthermore, we are using teacher-student method to aid the network in the compressed domain to mimic the teacher network in the raw domain. We show experiments on three leading datasets (HMDB51, UCF1, and Kinetics) that approach state-of-the-art accuracy on raw video data by using compressed data. Our model MFCD-Net outperforms prior methods in the compressed domain and more importantly, our model has 11X fewer parameters and 3X fewer Flops, dramatically improving the efficiency of video recognition inference. This approach enables applying neural networks exclusively in the compressed domain without compromising accuracy while accelerating performance.
\end{abstract}


\begin{figure*}
  \centering
 \includegraphics[width=0.9\linewidth]{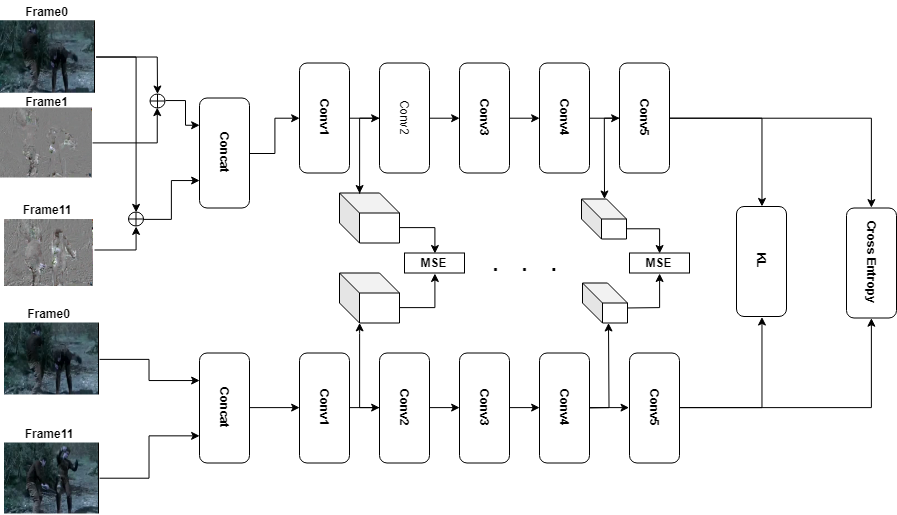}
  \caption{Overview of the proposed method showing the student-teacher framework where the student network is the network that is fed with a compressed clip which means 1\iframe and N augmented residual frames (N=11 in the figure) which can be seen in the upper part of the figure, below is the teacher network which is a pretrained network in the raw domain, we enforcing the compressed domain network $\Phi_{student}$ to mimic $\Phi_{teacher}$ by imposing $\Phi_{student}$'s intermediate feature maps to be similar to those of $\Phi_{teacher}$, by using MSE loss and KL loss on the logits}
  \label{fig:Student_Teacher}
\end{figure*}

\section{Introduction}
Video traffic is projected to capture 82\% of all Internet traffic by 2022 \citep{Cisco2019}. Video understanding with deep neural networks portends to be an important future workload. However, the spatiotemporal redundancies intrinsic in videos complicate the deployment of such models. State-of-the-art approaches are typically two-stream convolutional neural networks \citep{Simonyan2014} that require expensive computations of optical flow. Some works \citep{Zhang2018bd} tried to save the optic flow computation by using the motion vectors from the compressed stream as a coarse approximation of optic flow.

Video signals have significant redundancy in the spatial and temporal axis providing significant compression.
This redundancy makes video inference and computation a very expensive task in deep learning. The goal of this paper is to find a more efficient way infer actions from videos with a minimal decrease in accuracy.

Performing deep learning on video tasks in the compressed domain is a new task presented by Wu \etal \cite{Wu2018i}.
Inference in the compressed domain can save enormous amounts of storage, needed to store full RGB video clip and still infer with high accuracy.
Furthermore, it would much more efficient to do the computation directly on the compressed \textbf{bit} stream; this would save capacity, computation and more importantly, the video decoding time and hardware.
This is a very difficult task.
Our domain as in previous works \cite{Shou2019, Wu2018i, Huo2019} is parallel to working in the partial compressed domain which will be named simply the compressed domain.
Although our work is not on the compressed bit stream, this does take a step forward to the direction of working on a more compact representation.

In the compressed domain, there are three different components \iframes, residual frames and motion frames (motion vectors and residuals extracted from frames named \pframes). Recent attempts to accelerate such models in the compressed domain  \citep{Shou2019, Wu2018i, Huo2019} used the same idea of modeling three different networks, one for each modality mentioned above. Their work still has a significant gap from the performance achieved by operating on the uncompressed RGB frames. In addition, these approaches require employing multiple 2D CNNs for various information components.

We approach this task differently than previous research. Instead of thinking of the residual and the motion vectors as different modalities, we claim that the residual frames are able to replace the missing RGB frames using a few adjustments in the video clip. We leverage the Multi-Fiber Net \citep{Chen2018} as a 3D convolutional backbone and treat the video clip the same as a video in the raw domain,  which holds N RGB frames. In the compressed domain we have one \iframe and N-1 \pframes and we will use those N-1 \pframes as a replacement for the missing RGB frames, this clip will be processed together using a 3D CNN. Our method brings dramatic reductions in memory and computational complexity while increasing accuracy. Compared to previous works in the compressed domain \citep{Wu2018i, Shou2019}, our method uses one 3D CNN and teacher-student framework in order to achieve the state of the art performance. This way of interfacing with the task help our network in the compressed domain, MFCD-Net to be more similar to the network in the raw domain MFNet\citep{Chen2018}. 

\begin{figure*}
  \centering
  \includegraphics[width=0.7\linewidth,height=1.6in]{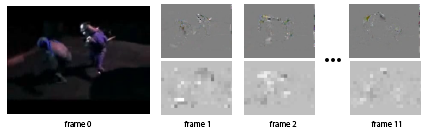}
  \caption{Example of a GOP structure: the \iframe is on the left, on the upper row  we see the residuals component, on the bottom are the motion vectors frames. The goal of this figure is to build the understanding of the compressed representation and to realize the small amount of data the residual frame consists.}
  \label{fig:residual_frame2}
\end{figure*}
We summarize our contributions as follows:
\begin{itemize}
    \item We introduce a novel approach to accelerate the inference of action recognition in the compressed domain. Our method looks at the compressed components as a single unit, hence we are able to use a single 3D CNN.
    
    \item We approach the compressed stream as a pseudo video stream helping us to train the network using a student-teacher framework, which for the first time being done between the compressed domain and the uncompressed domain.
     
    \item Unlike previous work, our approach uses only the \iframe and the residuals but not the motion vectors. This leads to memory capacity and bandwidth decrease and computation savings.

    \item Compared to previous approaches in the compressed domain \citep{Shou2019, Wu2018i, Huo2019}, our MFCD-Net beats the state of the art by 4.1 points on HMDB51\citep{Kuehne2011} top1 and 2.3 point on UCF101 \citep{KhurramSoomro2012} top1, while requiring $11\times$ fewer parameters and $3\times$ fewer FLOPs.
    
    \item Our approach leads to a minimal decrease in accuracy across two leading datasets: UFC101 \citep{KhurramSoomro2012} and HMDB51 \citep{Kuehne2011}. For example, our decrease in accuracy is only 0.9 on the UCF101 dataset.
    
\end{itemize}

\section{Related Work}
\paragraph{Video Action Recognition}
Action recognition in videos is a well-studied task with a variety of datasets \citep{KhurramSoomro2012, Kuehne2011, Kay2017}. The most common approach to recognition on videos with RGB frames is a two-stream approach \citep{Simonyan2014}, which applies convolutional neural networks to separately process RGB frames and optical flow, this method showed significant improvement over previous methods, but holds massive calculations by needing to compute two different networks and by creating optic flow.
The individual CNNs have recently used 3D convolution \citep{Carreira2017} to achieve state-of-the-art results. While 3D convolutions can exploit the temporal domain, prior work still uses optical flow to achieve higher results. Alternative approaches have applied RNNs \citep{Donahue2017} or other aggregation approaches to reason over the temporal domain.

While providing a significant performance increase, and having a widespread use in action recognition, its high parameter count and computational cost is an important disadvantage. Recently, the efficiency of the operator has been substantially improved by decomposing the 3D convolution filters \citep{Tran2018b, Xie2018}, or employing group convolutions such as in the Multi-Fiber Network \citep{Chen2018}. We used Multi-Fiber Network to demonstrate state of the art accuracy without necessarily requiring high computational cost.

\paragraph{Video Compression}
Video compression is the process of converting digital video into a format suitable for transmission or storage and reduces the number of bits by removing spatial and temporal redundancies. Video codecs splits the video into Group Of Pictures (GOP). The GOP consists of an Intra frame or \iframe, which is a self-contained RGB frame with full visual representation, followed by a sequence of inter frames (\pframes or \bframes) that only represent the changes that occur with respect to the previous frame. Here, we used a GOP with 12 frames (1 \iframe and 11 \pframes that represent the change in the frames) in order to compare our results with previous works \citep{Shou2019, Wu2018i, Huo2019}.

The \pframes hold two types of data: motion vectors and residuals as depicted in Fig.~\ref{fig:residual_frame2}),  Motion vectors represent the movement of a block of pixels from a source frame to a target frame. In the encoding phase, we predict the motion vectors, and use them in order to warp the source frame. 
The residual is calculated by subtracting the new frame resulting from the warping and the original frame.
The residuals represent the difference between the warped image using the predicted motion vectors and the target frame. In other words, the residual frame holds the error of the motion prediction. For simplicity, we will not use \bframes.

\begin{table*}[]
    \caption{Comparison with previous works: our model outperforms prior work in the compressed domain while requiring significantly fewer parameters and Flops.}
    \label{tab:results-table}
    \centering
\begin{tabular}{lrr|r|r}
      \toprule
       &      &      & {UCF-101} & {HMDB51} \\
       Method & Params.[M] & GFLOPS & Top-1  & Top-1 \\
      \toprule
CoViaR \citep{Wu2018i} & 83.6 & 3,615 & 90.4 & 59.1 \\
DMC-net (Resnet18) \citep{Shou2019}& 95.2 & 401 & 90.9 & 62.8 \\
TTP \citep{2019arXiv190810155H}& 17.5 & 1050 & 87.2 & 58.2 \\
      \textbf{MFCD-Net (Ours)} & \textbf{8.0} & \textbf{128} & \textbf{93.2} &  \textbf{66.9}  \\
      \bottomrule
\end{tabular}
\end{table*}

\paragraph{Action Recognition and the Compressed Domain}
Wu \etal \citep{Wu2018i} first proposed to apply deep learning for action recognition directly in the compressed domain with CoViAR. In order to build a frame based model, they used three different 2D CNNs, with each network trained on different types of compressed data: the \iframe, and the residuals and motion vectors of the \pframe. To boost performance, they also incorporated optical flow, which necessitated a full decoding of the video in order to extract the desired optic flow, which defeats one of the advantages of operating directly in the compressed domain; also it's adding a large capacity load and computation delay as each opticflow frame compute time is between 20-80ms. Other approaches \citep{Zhang2016a, Zhang2018bd} leverage optical flow and motion vectors, which are correlated, to transfer knowledge learned in optical flow.

Specifically, DMC-Net \citep{Shou2019} improved on CoViAR by introducing a lightweight generator to reduce motion vector noise and capture fine motion details. This approach does not require calculating optical flow, significantly reducing the computation time. DMC-Net, as Coviar, uses several 2D CNN models each for different modality.
Shou \etal \citep{Shou2019} showed accuracy results replacing their Resnet18 model with the Inflated 3D CNN model from Carreira and Zisserman \citep{Carreira2017}, but that model requires 250 frames at inference time (approximately 8 seconds with FPS of 30), which contradicts our goal of low latency inference and the requirement to perform inference on data in the compressed domain, hence we will not refer to this model.
Recently, Huo \etal \citep{Huo2019} followed Coviar \cite{Wu2018i} and DMC-Net \cite{Shou2019} and used also a few 2D CNNs each for every modality, for each 2D CNN they used Mobilenet \citep{Sandler2018} in order to be more efficient, they introduced a new block named Temporal Trilinear Pooling for combining the three modalities (I-frame, Motion Vectors, Residuals).
In summary, all previous works used the same idea with different tweaks; our method is unique and novel in the compressed stream, as explained in the next section.


\section{Method}\label{Method}
We desire to bring the inference in the compressed domain to a familiar place, the raw video domain; however, instead of having $N$ RGB frames we have only one \iframe, the remaining $N-1$ frames are \pframes.
Our Hypothesis is that the residual component in the \pframes can fill the role of the missing RGB frames, and we can process this clip which consist of one \iframe and $N-1$ residual frames as one unit using a 3D CNN, the path we are walking in order to achieve it, is described in the next few sections.

\label{headings}
\subsection{Uncompressed Domain}
We first present the notations of the network in the uncompressed domain and this network has two vital roles in our method:
\begin{itemize}
    \item Measure performance on the uncompressed RGB in order to act as a benchmark.
    \item  A teacher network to the network in the compressed domain (e.g MFCD-Net); it will be denoted as $\Psi_{teacher}$. 
\end{itemize}
We represent each video as a sequence of $K$ frames: $\mathcal{V_{RGB}}= [X_1, X_2, \ldots, X_K ]$, where each frame is an RGB image $X_{k} \in \mathbb{R}^{H\times W \times C}$. Therefore, the input video will have shape $(N, C, K, H, W)$, where $N$ is the batch size, $C$ number of channels and $H,W$ the spatial size of each frame.

The raw data is fed into $\Psi_{teacher}$, and can be written:
\begin{equation} \label{eq:1}
y_{t} =  \Psi_{teacher}(\mathcal{V_{RGB}}) 
\end{equation}
Where $y_{t}$ is the label predicted, $t$ represents the fact that this is the output of the teacher network.

\subsection{Compressed domain} \label{compressed}
\paragraph{Data modeling} Feeding the data into the network is one of the key obstacles in the analysis of videos in the compressed domain.

In the uncompressed domain, every frame is independent, since the content of the $k^{th}$ RGB frame is independent of the $k-1$ RGB frame. However, in the compressed domain, dependencies exist between the frames due to the use of temporal information. Each \pframe holds data with respect to the previous frame, which is likely another \pframe.

The goal of the data modeling is to make the \pframes independent, so we need to make a straight connection between the \iframe and each of the \pframes. In order to do that, we need to understand which pixels have moved and where.

We will need to trace the motion vectors back to the full image \iframe. We are modeling the compressed data similar to CoViAR \citep{Wu2018i} to remove the dependency on the \iframe.

\iffalse
We accumulate the residuals and trace the motion vectors back to the reference \iframe. Given a GOP with frames across 12 time steps of $t \in [0, 11]$, \iframe is at $t=0$. We denote a pixel in a \pframe at time $t$, as $n_p = (n_{p,x},n_{p,y})$.
Our goal is to connect $n$ to a pixel $n_i = (n_{i,x},n_{i,y})$, in the \iframe at time $k$, where $k<t$.

We can connect frame $t$ to frame $t-1$ using the motion vectors, denoted as $\Delta_{n_p}^t$, and the reference location in the previous frame will be:
\begin{equation} \label{eq:2}
\mu^{(t)}(n_p) = n_p - \Delta_{n_p}^t 
\end{equation}

In order to establish the connection between the \pframe at time $t$ and a frame at time $k$, we can denote:

\begin{equation} \label{eq:3}
\Gamma_i^{t,k} =  \mu^{(k+1)}(n_p) \circ \mu^{(k+2)}(n_p) \circ \cdots \circ \mu^{(t)}(n_p)
\end{equation}

If we assume that the $k^{th}$ frame is the \iframe in the GOP, means $k=0$, we can write:
\begin{equation} \label{eq:4}
\Gamma_{n_p}^{t,k} =  n_i
\end{equation}
Finally the accumulated motion vectors can be noted as:
\begin{equation} \label{eq:5}
D_{n_{p}}^{t,k} = {n_p} - \Gamma_{n_p}^{t,k}
\end{equation}
The accumulated residual frame is:
\fi
\begin{figure*}
  \centering
  \includegraphics[width=0.8\linewidth,height=2.1in]{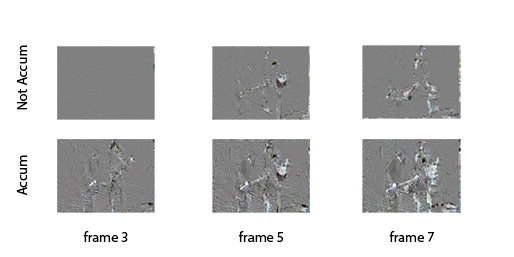}
  \caption{An example of the difference between residuals and accumulated residuals. The accumulate method holds the information through the time steps. Frame 5 shows the effect of previous frames (unlike non-accumulating method), thus we exploit the temporal data.}
  \label{fig:accumulated_vs_regular}
\end{figure*}
We will summarize the data modeling using our notation. Given a GOP that consists $K$ frames,where the \iframe is frame number $0$ and the rest (\ie $[1, K-1]$) are \pframes.
We'll denote a pixel in a \pframe as $n_p = (n_{p,x},n_{p,y})$, and a pixel in a \iframe as :$n_i = (n_{i,x},n_{i,y})$.
Our goal is to connect a pixel in a \pframe at time $t$ to a pixel in the \iframe $n_i$, in the \iframe at time $k$, where $k<t$.

The pixel $n_p$ in  the residual frame at time $k \in [1,K-1]$ is denoted as $R_{n_p}^k$ and the accumulated residual frame from time step 0 (i.e \iframe) to time step k is denoted as $\bar R_{n_p}^k$.
Although we are not using the motion vectors in our method, it is needed to explain the data modeling. The accumulated motion vectors frame from \iframe to time step $t$ is denoted as $\bar D_{n_p}^{k}$.
Now, we have a direct connection between each \pframe and the \iframe, can be written as:
\begin{equation} \label{eq:7}
U_{n_p}^k = U_{{n_p}-\bar D_{n_p}^{k}}^{0} + \bar R_{n_p}^k 
\end{equation}
Where $U_{n_p}^{k}$ is the $n_p$ pixel in the RGB frame at $t=0$.

Intuitively, equation \ref{eq:7} shows the direct connection between the \pframe at time step $t$ to the \iframe, by looking a specific pixel in the \iframe, moving it to its place in a \pframe at time $t$ and adding the accumulated residuals.

In Figure \ref{fig:accumulated_vs_regular}, we're able to see the advantages of this method. At time step 3, the residual has almost no data, while the accumulated residual accumulate the new data upon the previous data and give this small amount of data context.
Furthermore, we can see time step 5, the residual frame mainly shows one figure in the frame while the accumulated residual shows two which again helps the 3D CNN to understand the full context of each frame.

\subsection{Model} 

In order to propagate features from the \iframe to the rest of the residual data in the GOP, we introduce a simple but effective method of accumulated residuals across the time steps. We evaluated several merging techniques and found that the most efficient and accurate method is to simply add the \iframe to each residual frame. This helps preserve the change that is expressed by the residual frames but still propagates the scene background, color and texture. Given an \iframe, denoted as $U^{0} \in R^{HxWx3}$, and accumulated residual frames $\bar R^t \in R^{HxWx3}$ for each time step $k$, we can write the augmented residuals $\hat R^k$ as:
\begin{equation} \label{eq:8}
 \hat R^k = \frac{(\bar R^k+U^{0})}{2},    \forall k \in [1,K-1]
\end{equation}
{
\begin{prop} The residual frames (i.e $\hat R$) holds enough information to replace the original RGB frames from the original video.
\end{prop}
Once we follow the above proposition, we can use the \iframe and the residual frames as one video clip, where instead of 12 RGB frames, the clip holds one \iframe and 11 residuals.This enables the use of a 3d convolution network, which showed large improvement in video understanding tasks.
Hence our video clip will look as follows: 
\begin{equation} \label{eq:9}
y_s =  \Psi_{student}([U^{0},\hat R^{1},...,\hat R^{K-1}])
\end{equation}
Where $s$ is to mark that $y_s$ is the prediction for the student network.
The augmented residuals operation (i.e adding \iframe to each residual frame) is depicted in Fig.~\ref{fig:compress_add}.
}
\paragraph{Student Teacher} So far we discussed the network in the compressed domain (i.e MFCD-Net) with the data modeling process and the way of looking at the task and data modeling them self bring to state of the art results. However, we desire to take our model a step forward, in order to aid our network in the compressed domain to learn beneficial features that are similar to those in the raw domain (which we desire to be similar to), therefore we use student-teacher framework \citep{Romero2015}\citep{Hinton2015}.
In this framework the output of a teacher’s hidden layers and logits are responsible for guiding the student network to predict similar feature maps.

Our notations are as follows: $\Psi_{student}^i$ will represent the the $i$'th layer in the student network, (i.e the network in the compressed domain), $\Psi_{teacher}^i$ will represent the the $i$'th layer in the teacher network, (i.e the network in the raw domain).
Our new objective function has three ingredients:
\begin{enumerate}
\item Hint sharing between the hidden layers, using $L_2$ distance as presented by Romero \etal \citep{Romero2015}, denoted as:
\begin{equation}
L_{hints}^i = \frac{1}N\sum_{n=1}^{N}{||\Psi_{student}^i - \Psi_{teacher}^i||^2}
\end{equation}
This objective will force the intermediate feature maps of the student network to look like the intermediate feature maps of the teacher network.
\item KL distance between the soft logits of the two networks as presented by Hinton \etal \citep{Hinton2015}. Let's first denote the soft logits of the student network as: $SL_{student}=Softmax(y_n^s/\tau)$
and of the teacher network:
$SL_{teacher}=Softmax(y_n^t/\tau)$
this can be denoted as:
\begin{equation}
L_{SL} = KL(SL_{student}||SL_{teacher})
\end{equation}
Where KL() is the Kullback-Leibler distance.

\item Cross entropy loss between the predicted output and the label, denoted as:
\begin{equation}
L_{CE} = CE(y^s,\tilde{y})
\end{equation}
Where $\tilde{y}$ is the ground-truth label and $y^s$ is the predicted label.
\end{enumerate}

The total objective loss is
\begin{equation}
L_{total} = L_{ce} + \lambda_1\sum_{i=1}^{h} L_{hints}^i 
+ \lambda_2*L_{SL}
\end{equation}
Where $\lambda_i$ indicates the effect of each loss on the total loss function and h is the number of hidden layers.
Our full method is depicted in Fig.~\ref{fig:Student_Teacher}.

\begin{table*}[]
  \caption{Performance of our full  method  using residual augmentation and Student Teacher learning compare with the MFNet in the raw domain. This is video level accuracy with 12 frames for each clip, and an average of the 3 splits}
  \label{classification-performance-table}
  \centering
\begin{tabular}{lrrr|rrr}
\toprule
 & \multicolumn{3}{c}{HMDB51} & \multicolumn{3}{c}{UCF-101}\\
 & RGB & \emph{Ours} & $\Delta$ & RGB & \emph{Ours} & $\Delta$\\
\toprule
MFCD-Net (Ours)  & $69.6$ & $66.9$  & 2.7   & $94.1 $ & $93.2$  & 0.9  \\
\bottomrule
\end{tabular}
\end{table*}

\section{Ablation Study}
In order to understand what is the contribution of each part in our model, we also tested the model's performance when trained on formats different from the previous section. 
\paragraph{No Transfer Learning}
In this experiment, we are not training using the student-teacher (Transfer Learning) framework in order to understand the benefit of this method.
A visualization of this experiment can be seen at figure \ref{fig:compress_add}.

\begin{figure}[h]
  \centering
  \includegraphics[width=0.45\textwidth]{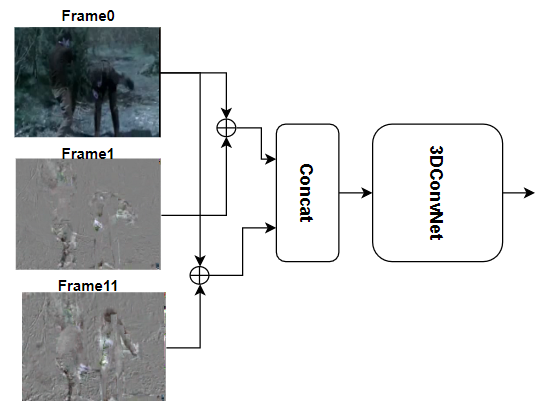}
  \caption{This Figure shows the idea behind adding the \iframe to the rest of the residual frames in the GOP}
  \label{fig:compress_add}
\end{figure}

\paragraph{No Augmentation of Residuals.} Although the residual frames have little information, we hypothesize that they contain sufficient information for an accurate prediction, without requiring augmentation of the residuals with the \iframe, as used in the previous section. We write this as:
\begin{equation} \label{eq:10}
y =  \Psi([U^{0},\bar R^1,...,\bar R^{11}])
\end{equation}
The model can use the solo \iframe at the beginning of the GOP as an ``anchor'' to extract features such as color and texture, that are then implicitly propagated to the residual frames. Figure \ref{fig:compress_no_add} shows an example of the input GOP, 1 \iframe and 11 residual frames.

\paragraph{Residuals Only} We also examine the challenging case where our model only has access to the accumulated residuals $\bar R$. We can write this method as
\begin{equation} \label{eq:11}
y =  \Psi([\bar R^1,\bar R^1, \bar R^2, ...,\bar R^{11}])
\end{equation}
In order to keep the number of input frames to 12, we have duplicated the first residual frame once. While this case will never occur in the compressed format of data, its an important ablation study to better understand how 3D CNNs operate in the compressed domain. 

\section{Experiments}
In this section, we discuss the implementation details of our experiments, datasets used, and results achieved.
\paragraph{Datasets}
We report results on three leading datasets, including HMDB-51\citep{Kuehne2011},
UCF-101\citep{KhurramSoomro2012}, and Kinetics400 \citep{Kay2017}.
HMDB-51 contains 6,766 videos from 51 action categories, while UCF-101 contains 13,320 videos from 101 action categories. Both datasets have 3 officially specified training/test splits. 
Our result are calculated as an average on the three splits.
Kinetics dataset holds 400, with 400\textemdash1150 clips for each action. The original version has 306,245 videos, and is divided into three sets, training, validation and test.
As Kinetics is a list of videos from YouTube, some videos were deleted, hence today out dataset holds around 200k videos for training and 40k videos for validation. Our experiments are done on the validation set.

\paragraph{Training} The models were pretrained on Kinetics in the raw domain, then fine-tuned for our task in the compressed domain using cross-entropy loss and SGD optimizer, with a learning rate of 0.005 along with weight decay of 0.0001 and 0.9 for momentum. During training, we split the video into 12-frame clips in order to be aligned with previous works and use the data preparation procedures outlines in the section ~\ref{Method}. For the MFNet3D network, each frame in the clip was resized to $256\times 256$ and cropped to a $224\times 224$ frame and horizontally flipped with a 50\% probability.

\paragraph{Inference} During inference we randomly sampled fifteen 12-frame clips (i.e 1 \iframe and 11 residual frame) to generate input clips for our networks, and each clip is randomly cropped. 
Each clip pass requires 8.53 GFLOPS; therefore, in order to do one prediction we will require 128 GFLOPS.

Our advantage over previous work is the fact that we are seeing the residuals as a substitute of the missing RGBs and not a different modality, which allows us to use only one network and not multiple 2D CNNs.


Our experiment section has four components:
\begin{itemize}
    \item Compare the accuracy of our full method (i.e MFCD-Net) with the RGB raw domain accuracy using MFNet. 
    \item Compare our MFCD-Net with other approaches in the compressed domain.
    \item Extensive ablation study.
    \item Kinetics experiments comparison with raw domain networks.
\end{itemize}

\subsection{Accuracy comparison: compressed \vs uncompressed domain}
We applied our method on MultiFiberNet3D \citep{Chen2018}. The model was pre-trained on the Kinetics \citep{Kay2017} raw data. Then it was fine tuned on leading datasets: HMDB51 \citep{Kuehne2011} and UCF101 \citep{KhurramSoomro2012} raw data in order to be our reference benchmark. The same pre-trained models were also fine tuned on HMDB51, UCF101 and Kinetics compressed data representation (as described in Section ~\ref{compressed}).

As shown in Table \ref{classification-performance-table}, the difference in performance between using the raw RGB videos and the video representation in the compressed domain are small, and our method takes a significant step towards making the work in the compressed domain more practical, showing less than 1 top1 decrease in UCF101 dataset and a decrease of 2.7 points in accuracy in HMDB51.

\begin{figure}[h]
  \centering
  \includegraphics[width=0.45\textwidth]{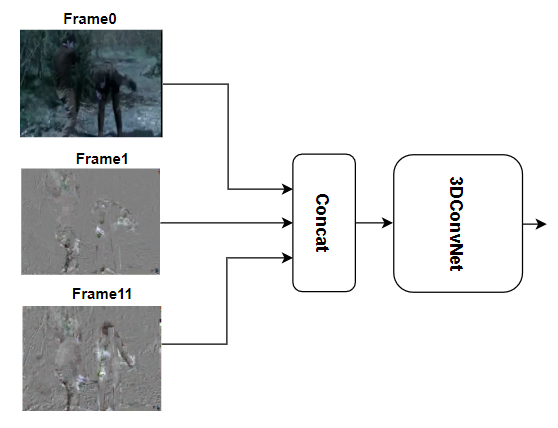}
  \caption{This figure shows the basic configuration of feeding a group of picture in the compressed domain into a 3DCNN}
  \label{fig:compress_no_add}
\end{figure} 

\subsection{Comparison with State-of-the-Art}
In the field of Action Recognition in the compressed domain, there are three leading models: CoViAR \cite{Wu2018i}, DMC-Net \citep{Shou2019} and TTP \cite{Huo2019}. We compare our model in terms of accuracy (Top-1), computational requirements (GFLOPS) and memory requirements (number of parameters). 
Coviar \citep{Wu2018i} test procedure is as follows: from each video they sampled 25 \iframes, 25 residual frames, and 25 motion vector frames, with each frame is getting 2 different flips and 5 different crops for data augmentation \cite{Huo2019}, and followed the settings of Wu \etal \cite{Wu2018i}.
DMC-Net \citep{Shou2019} partially followed Coviar settings \citep{Wu2018i}, except the data augmentation part.
Those important details must be considered when attempting to solve this task, as efficiency is a major criterion.

\paragraph{Computation Evaluation} Regarding computational requirements in video action recognition, we need to examine how many GFLOPS are needed for one prediction (which consist of multiple data passes). Previous works presented the average GFLOPS required for processing each frame; even if we presented the GFLOPS as previous works did, per frame, our method consume 8.53 GFLOPS for each clip hence each frame is $8.53/12 = 0.71$ GFLOPS (which is the most efficient method).
But calculating GFLOPS per frame losses any correlation to the accuracy reported, since it depends on the test procedure, hence we will report the GFLOPS required to infer 1 video. Our method samples 15 clips from each video, then our total amount of GFLOPS is: $8.53*15 = 128$.
For example, TTP \cite{Huo2019} claims to require 1.4 GFLOPS, which is the average number of required operations for passing 1 frame through MobileNetV2 \citep{Sandler2018}, but their inference procedure is the same as Coviar, which means that they are sampling 75 frames and for each frame they are doing 10 augmentations, so there are 750 passes through Mobilenet, which sums up to $750*1.4 = 1050$ for a single video inference.


As shown in Table \ref{tab:results-table}, our model significantly outperforms the previous state-of-the-art. In HMDB51, our accuracy is higher by 4.1 points from the current state-of-the-art, 7.8 from Coviar and 8.7 top1 points from TTP. In UCF101, our accuracy is higher by 2.3 top1 points from the state-of-the-art DMC-Net, 2.7 from CoViar and 6 top1 points higher than TTP. With the impressive leap in accuracy mentioned above, it's expected that requirements should be massive but our model is more efficient than previous work as we no longer need to keep three or four 2D CNNs, we no longer need heavy augmentation process and our departure from the motion vectors helps us with efficiency. Our experiments shows that our method requires 68\% less GFLOPS and 92\% reduction in memory footprint than the state-of-the-art DMC-Net. Model size is critical for performance, as it allows the model to be cached on an accelerator's local memory, leading to a significant increase in performance.

\subsection{Ablation Study Results}\label{ablation}
In this section, we discuss the importance of each and every element of our method, using the results from the experiments presented in Table \ref{tab:ablation}. We see that the student-teacher framework increased the performance drastically, which helps us to get closer to the raw domain. We also say that the Augmented Residuals phase, where we add \iframe to the residual frames in the GOP holds limited contribution, if any, to top1 accuracy.
When looking at the last row we can understand how important is the first and only \iframe in the compressed clip; more than 4 top1 points decreased when we dropped the \iframe from the clip.
\begin{table}
\caption{Performance for our ablation study as mentioned in Section \ref{ablation}: for HMDB51 and UCF101 datasets, performance was measured on split1.}
  \label{tab:ablation}
  \begin{center}
  \begin{tabular}{llrr}
    \toprule
 Model & Format &  HMDB51 &  UCF101   \\
    \midrule
 MFCD-Net & Full Method  & 66.9  & 93.2 \\
 MFCD-Net & \footnotesize I + Agmt. Residuals $\hat R$ & 64.8  & 92.4 \\
 MFCD-Net & I + Residuals $\bar R$ &  64.4 & 92.5 \\
 MFCD-Net & Residuals only $\bar R$ & 60.1   & 89.0 \\
    \bottomrule
  \end{tabular}
  \end{center}
\end{table}
\subsection{Kinetics}
We trained our model on the Kinetics400 training set (raw domain). We then fine-tuned our network on Kinetics (compressed domain) and evaluated on Kinetics validation set (compressed domain). We compared our results to models presented in \citep{Hara2018}, evaluated on the Kinetics validation set. Those models were trained and evaluated in the raw domain. Our model significantly beats those models.

\section{Discussion}
{
One might claim that our use with 3dCNN made our results much better, that is absolutely correct, Our contribution was in suggesting the hypothesis, which residual frames can replace RGB frame and suggesting a piratical way of doing it, this novelty enabled us to use 3dCNNs which are much more suitable for videos instead of 2dCNNs as used in previous works.
Video is a redundant data type. It effects our ability to analyze this data type as it requires us massive calculations and storage, with little ROI. One of the ways to reduce this redundancy and reduce storage massively is to work in the compressed domain. As shown above, video compression is the best redundancy exploiter that can be found.
The primary savings by using residual frames comes from their small memory footprint. Further work should focus on reducing the model size, since now the video consists less data than the RGB video clip.

}

\section{Conclusions}
In this paper, we introduce a novel way to examine the task of action recognition in the compressed domain. Different from previous research, we claim that the components of the compressed stream can be seen as a unified entity and that the residual frames are able to supplant the omitted RGB frames and form a pseudo raw video clip along with the \iframe at the beginning of the GOP. Our method did not use motion vectors from the compressed stream (in order to save storage and computation), nevertheless we achieved state-of-the-art results beating previous research \citep{Wu2018i, Shou2019, Huo2019} by a large gap, with dramatically fewer parameters and lower computation requirements. Furthermore, we achieved minimal degradation in performance compared to the same architecture RGB frames \citep{Hara2018, Chen2018}, which is the main purpose of this research (\ie to achieve the accuracy of a network in the raw domain working in the compressed domain).

Although the field of action recognition in the compressed domain has a few practical issues, it holds great potential to more efficient inference and training. Our method and experiments are a step forward towards enabling efficient deployment of video recognition in all platforms.
\begin{table}
  \caption{Top-1 accuracy on Kinetics val. set: results for other models are from Table 2 of Hara \etal \citep{Hara2018} and are using the uncompressed RGB data; our model uses the compressed representation.} 
  \label{tab:kinetics}
  \centering
  \begin{tabular}{lr}
 \toprule
 \cmidrule(r){1-2}
 Model  & Kinetics Top-1 \\
  \midrule
 DenseNet-201 & 61.3 \\
 ResNet-152 & 63.0 \\
 ResNet-200 & 63.1 \\
 Wide ResNet-50 & 64.1 \\
 ResNext-101 & 65.1 \\
 MFCD-Net (Ours) & \textbf{68.3} \\
 \bottomrule
  \end{tabular}
\end{table}

\clearpage
{\small
\bibliographystyle{ieeetr}
\bibliography{ms}
}
\end{document}